\documentclass{ifacconf}

\usepackage{graphicx}      
\usepackage{natbib}        
\usepackage{amsmath}
\usepackage{amsfonts}
\usepackage{comment}
\usepackage{float}
\usepackage{todonotes}
\usepackage{siunitx}
\usepackage{subcaption}

\begin{document}
\begin{frontmatter}

\title{Sim2Swim: Zero-Shot Velocity Control for Agile AUV Maneuvering in 3 Minutes} 



\author{Lauritz Rismark Fosso,} 
\author{Herman Biørn Amundsen,}
\author{Marios Xanthidis,}
\author{Sveinung Johan Ohrem}

\address{SINTEF Ocean, Trondheim, Norway (E-mail: lauritz.fosso@sintef.no; herman.biorn.amundsen@sintef.no; marios.xanthidis@sintef.no; sveinung.ohrem@sintef.no)}  

\begin{abstract}                
Holonomic autonomous underwater vehicles (AUVs) have the hardware ability for agile maneuvering in both translational and rotational degrees of freedom (DOFs). 
However, due to challenges inherent to underwater vehicles, such as complex hydrostatics and hydrodynamics, parametric uncertainties, and frequent changes in dynamics due to payload changes, control is challenging. 
Performance typically relies on carefully tuned controllers targeting unique platform configurations, and a need for re-tuning for deployment under varying payloads and hydrodynamic conditions.
As a consequence, agile maneuvering with simultaneous tracking of time-varying references in both translational and rotational DOFs is rarely utilized in practice.
To the best of our knowledge, this paper presents the first general zero-shot sim2real deep reinforcement learning-based (DRL) velocity controller enabling path following and agile 6DOF maneuvering with a training duration of just 3 minutes. 
Sim2Swim, the proposed approach, inspired by state-of-the-art DRL-based position control, leverages domain randomization and massively parallelized training to converge to field-deployable control policies for AUVs of variable characteristics without post-processing or tuning.
Sim2Swim is extensively validated in pool trials for a variety of configurations, showcasing robust control for highly agile motions.





\end{abstract}


\end{frontmatter}

\section{Introduction}
Autonomous underwater vehicles (AUVs) are fundamental in many critical ocean operations, including resource utilization, marine archaeology, maritime safety, and infrastructure maintenance.
Industries, such as offshore wind farms and aquaculture, rely on continuous, resilient inspection and intervention from underwater robots~\citep{transeth2024safesub, khalid2022applications, kelasidi2023robotics},
while AUVs are used for environmental monitoring~\citep{fossum2019toward}, seabed mapping~\citep{Ludvigsen:2016:autonomous_mapping}, and archaeological surveys~\citep{diamanti2025marine}. 

\begin{figure}[t!]
    \centering
    
    \begin{subfigure}[t]{0.48\textwidth}
        \centering
        \includegraphics[width=\linewidth, trim={0in 0in 0in 0in}, clip]{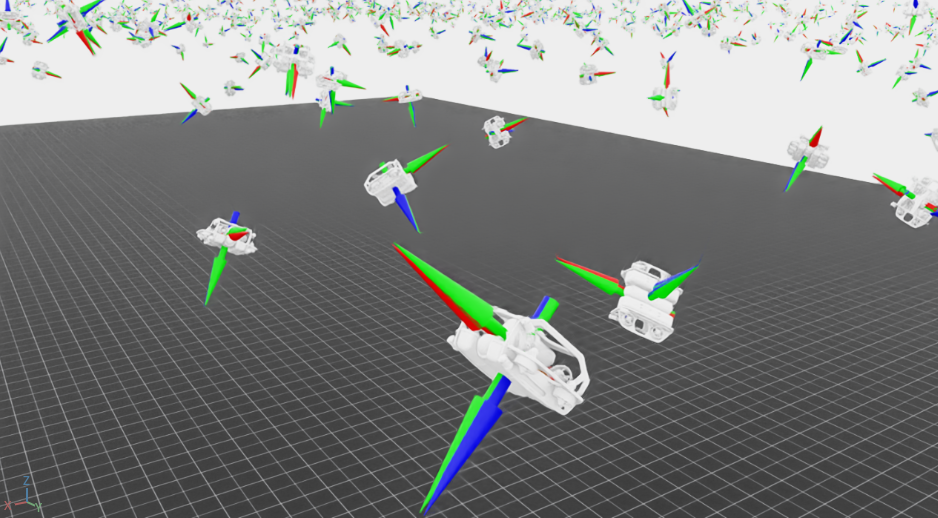}
        \caption{}
        \label{fig:bluerov}
    \end{subfigure}

    \vspace{0.5em}

    \begin{subfigure}[t]{0.237\textwidth}
        \centering
        \includegraphics[width=\linewidth, trim={20cm 20cm 20cm 20cm}, clip]{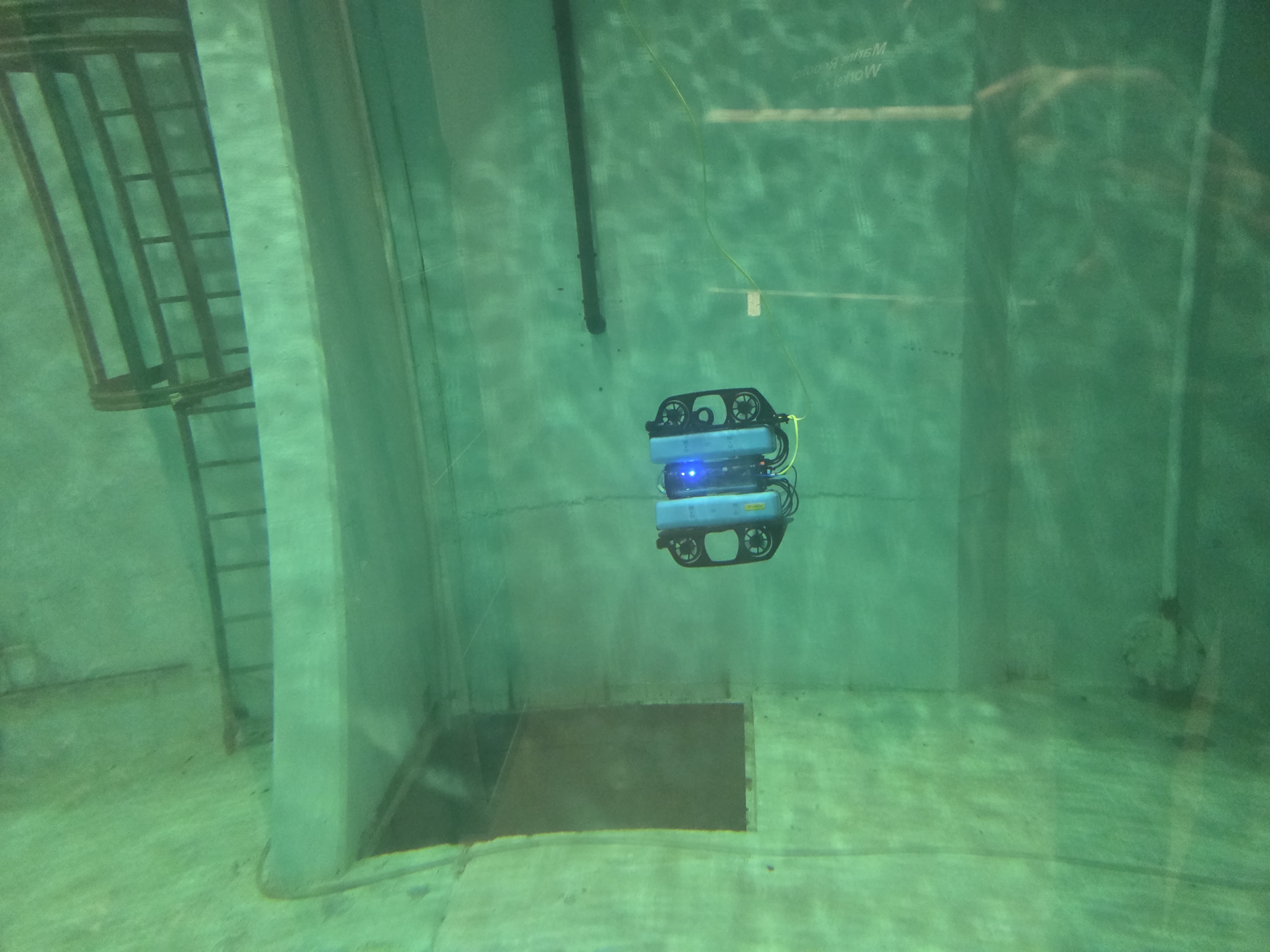}
        \caption{}
        \label{fig:murk:b}
    \end{subfigure}
    \begin{subfigure}[t]{0.237\textwidth}
        \centering
        \includegraphics[width=\linewidth, trim={20cm 10cm 20cm 30cm}, clip]{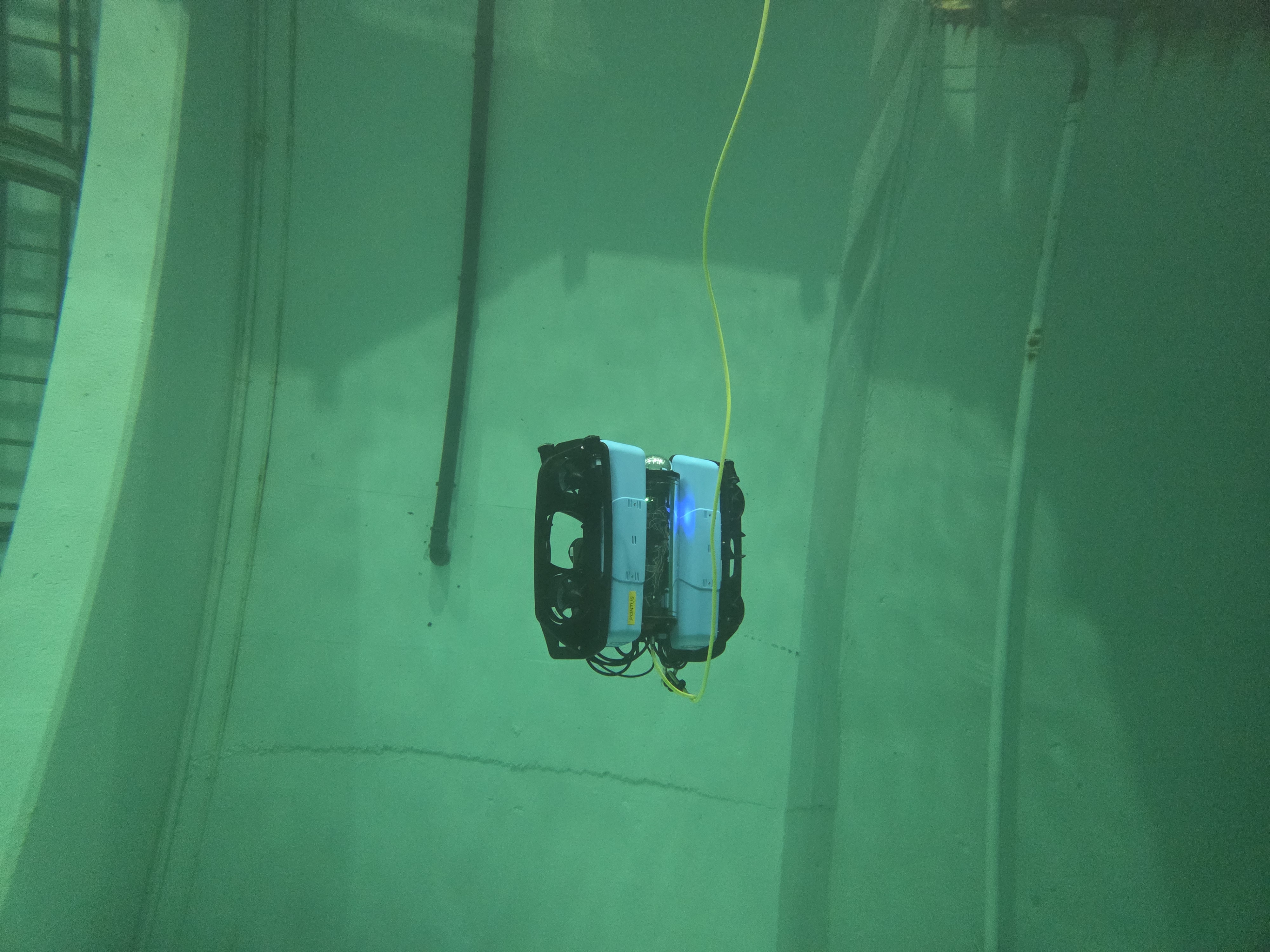}
        \caption{}
        \label{fig:murk:c}
    \end{subfigure}

    \caption{An instance of massively parallelized training with 2048 simulated robots in Isaac Sim is shown in (a). Deployment snapshots, in (b) and (c), showcasing robust complex maneuvering after 3 minutes of training.}
    \label{fig:beauty}
\end{figure}

All above operations require robust control in order to be able to execute motions to accomplish tasks, such as reaching a desired position and orientation, moving along a desired path at a desired speed and attitude, and performing contact interaction with the surroundings.
Path following, an integral capability of AUVs, is often achieved by utilizing a guidance law~\citep{Breivik:2005:CDC,caharija2016integral} to produce reference signals for the controller that steer the vehicle towards the desired path.
While underactuated vehicles rely on controlling the attitude of the vehicle to achieve path convergence, fully actuated vehicles can instead use linear velocity control to steer the vehicle towards the path~\citep{Breivik:2005:CDC}, thus leaving the orientation of the vehicle free to achieve some other mission independent of the path. 
Building upon the traditions from remotely operated vehicles (ROVs), fully actuated holonomic vehicles are popular for operations in proximity to infrastructure. 
They are built for precise, low-speed maneuvers, and have geometry that yields favorable stability properties, all while being modular platforms for variable payloads.
However, their dynamics are coupled and highly nonlinear, while parametric uncertainty is large due to the many appendages and cavities typical of such vehicles.
This make damping and added mass coefficients hard to identify, which is crucial when they are exposed in time-varying environmental disturbances from ocean waves and currents, especially for smaller vehicles. 
Commonly used hand-tuned PID controllers are vulnerable to these nonlinearities and changes in dynamics, often requiring fine-tuning of gains between deployments. 
Other control methodologies, such as sliding mode~\citep{Shtessel:2012:automatica:supertwisting} or adaptive control~\citep{ohrem_application_2024}, can be more robust to model uncertainties, but rely on expert knowledge and careful tuning.

Deciding on, developing, and tuning control approaches in targeted systems is highly logistically and effort-demanding, often requiring multiple deployments, work-hours, expertise, and instincts.
Even successful implementations of this process, which targets specific platforms, will result in suboptimal performance not only across deployments, due to variable loads and environmental conditions, but also during a single deployment in cases of interaction with the environment (e.g., lifting or picking objects, initiating contact with infrastructure, etc.).


To address all aforementioned challenges, in this work, we present \textit{Sim2Swim}, to the best of our knowledge, the first general zero-shot sim2real deep reinforcement learning pipeline for agile and robust 6DOF velocity and attitude control, Figure~\ref{fig:beauty}. 
Sim2Swim requires only trivial human input, and eliminates the logistics which come with user-driven controller parameter tuning, and demonstrates 6DOF maneuvering of high complexity after only 3 minutes training on a commercial-grade laptop.

The proposed methodology is inspired by and builds upon previous work~\citep{cai_learning_2025} that introduced massively parallelized training for position keeping.
It expands it by enabling agile path following through robust velocity control and eliminates previously reported steady-state errors by incorporating integral action --- all while achieving close to an order of magnitude faster convergence with less hardware requirements.
Our demonstrations showcase superior performance with agility in both translation and orientation in a variety of configurations with different payloads, 
providing new opportunities to fully utilize the hardware capacity of such platforms, such as for developing new agile policies employed for inspection of geometrically complex structures ~\citep{xanthidis_aquavis_2021}, with increased robustness to platform variability.  



In summary, the contributions of this paper are:
\begin{enumerate}
    \item Sim2Swim, a general reinforcement learning-based pipeline, trained in 3 minutes and converging in less than 2 minutes, for robust 6 DOF linear velocity and orientation control.
    \item An integral enhancement to eliminate steady-state errors, which enables resilient acrobatic behavior while accounting for payload and parameter changes.
    \item Extensive in-pool validation of the proposed pipeline with different configurations, showcasing superior zero-shot agility and stability.
\end{enumerate}

\section{Related Work}
The effectiveness of reinforcement learning (RL) has generated research interest in its potential control applications, and has shown promising results in control of quadcopters \citep{panerati_learning_2021,eschmann_learning_2024}, quadrupeds \citep{tsounis_deepgait_2020} and robot manipulators \citep{gu_deep_2017}. RL has also proven effective at handling complex tasks in other domains, such as manipulating deformable objects in the food industry, such as fish, which are soft and slippery \citep{herland_non-prehensile_2025}. Using RL to perform complex or precise tasks has been associated with long training times, but recent technological advances in GPU-based parallelized computing have enabled the development of highly parallelized RL. This motivated \cite{rudin_learning_2022} to develop Isaac Lab, a framework for RL for Isaac Sim~\citep{nvidia:2025:isaac_lab}. 
They exemplified its usage by training quadrupeds to walk in minutes.

 \cite{eschmann_learning_2024} further demonstrated the capabilities of GPU-based parallelization by developing a hyper-efficient simulator, which they used to train a policy to stabilize a nanocopter in 18 seconds. 
 They proposed an end-to-end approach, directly setting motor rotational speed setpoints.
 They argued that this enabled the policy to compensate for motor dynamics. 
 To achieve these training speeds, they used a training curriculum, and achieved a policy that was performant and robust to uncertainties, despite not employing domain randomization.

Inspired by the fast training speed enabled by Isaac Lab, and motivated by the frustration of having to constantly re-tune their controllers after changing their AUV's sensor configuration, \cite{cai_learning_2025} developed a custom simulation environment for AUVs in Isaac Lab.
Their ambition was to train a policy to perform setpoint regulation of position and attitude that was sufficiently performant, while robust to modeling uncertainties, in 15 to 20 minutes. 
To ensure robustness, they employed domain randomization.
Similar to \cite{eschmann_learning_2024}, they proposed a policy that directly sets each thruster's rotational speed.
While able to stabilize tracking errors, they report that steady-state errors were still present in sway and pitch.

Other works on DRL for underwater robotics includes~\cite{hadi_deep_2022}, who developed an integrated approach for both path planning and path following with obstacle avoidance for the REMUS 100 AUV by controlling its rudder. This approach however suffers from long (60 hours) training times, and has not been validated in field experiments.

A low-level actor-critic goal-oriented RL controller was developed and demonstrated by~\cite{carlucho2018adaptive} on a torpedo-shaped vehicle. 
The raw sensory information of the AUV was used as inputs to the RL architecture and the thruster commands as outputs. Experiments controlling the linear velocities (surge, sway, and heave) and transversal axes motions (yaw rate and pitch rate) show accurate tracking. The angular velocities were constrained to zero, though they were still controlled. 
However, no wall-clock training time is reported to allow comparative analysis.

To mitigate model uncertainties~\cite{ma_neural_2024} proposed a modification to the proximal policy optimization (PPO) scheme, called ModelPPO, to perform 3D path-following for underactuated AUVs controlling the course and elevation errors. This modification integrates a third neural network into the existing PPO architecture. This network represents the AUV model, which learns the state transitions of the AUV, and given the actions, outputs the predicted next state to the critic network. Their comparative simulation study showed a faster convergence over the PPO algorithm in unperturbed environments.

\cite{wang_imitation_2025} proposed imitation learning as an approach to perform path-following with a fully actuated AUV, with a comparative simulation study including the PPO algorithm. They argued that their imitation learning approach achieved similar results to PPO in less time. 

\cite{sufan_swim4real_2025} developed an approach to train an energy-efficient policy to perform 6-DOF setpoint regulation, which, after 15 hours of training, saw a 39\% decrease in energy consumption compared to a PID controller. Similar to~\cite{cai_learning_2025}, they employed domain randomization by varying the mass by 0.7\%. The approach is experimentally validated using trajectory tracking of constant setpoints (all setpoints are zero) in all DOFs.

Unlike previous approaches, we present a general agile zero-shot solution for 6DOF tracking with time-varying velocity and orientation references that requires less than 3 minutes of training.
We validate the method in laboratory experiments where the vehicle was able to perform path following by tracking linear velocity references generated by a 3D line-of-sight guidance law, while simultaneously tracking arbitrary orientation references.
The remainder of the paper expands on the details and our testing in the following sections. 
Section~\ref{sec:problem} formalizes the control problem, Section~\ref{sec:method} introduces the proposed DRL policy and massive parallelization framework, Section~\ref{sec:sim2real} presents results from zero-shot sim2real employment in a pool, before Section~\ref{sec:conclusion} concludes the paper.

\section{Problem Description}\label{sec:problem}
The objective of this work is to control velocity and attitude for holonomic AUVs capable of agile maneuvering.
The solution should be robust to parametric uncertainties caused by varying payloads and hydrodynamic conditions and be applicable in a number of scenarios, without tight integration to a guidance law or path planner. 

Formally, we first introduce the following reference frames:

--- \textit{North East Down (NED) Frame $\{n\}$:} This frame has its origin on the water surface, its $x$-axis pointing North, $y$-axis pointing East, and its $z$-axis pointing down. Vectors represented in this frame carry the superscript $n$.

--- \textit{Body Frame $\{b\}$:} This frame has its origin in the geometrical center of the vehicle, with its axis definitions following the SNAME convention. Vectors represented in this frame carry the superscript $b$.

Let $\boldsymbol{q}\in\mathbb{H}$ denote the unit quaternion that represents the attitude of $\{b\}$ relative to $\{n\}$, $\boldsymbol v^b = [u,v,w]\in\mathbb{R}^3$ the linear velocity of the vehicle,  $\boldsymbol{\omega}^b\in\mathbb{R}^3$ its angular velocity, while $\boldsymbol{\nu} = [\boldsymbol{v}^b, \boldsymbol{\omega}^b]$ denotes the combined linear and angular velocities.
Furthermore, consider time-varying reference signals $\boldsymbol{v}^b_d(t)\in\mathbb{R}^3,\boldsymbol{q}_d(t)\in\mathbb{H}$, 
with tracking errors
$\boldsymbol{v}^b_e(t)=\boldsymbol{v}^b-\boldsymbol{v}^b_d(t)$ and $\boldsymbol{q}_e(t)=\bar{\boldsymbol{q}}_d(t)\boldsymbol{q}$, where $\bar{\boldsymbol{q}}_d(t)$ denotes the conjugate of $\boldsymbol{q}_d(t)$.
The AUV is governed by the equations~\citep{fossen_handbook_2021} shown below:
\begin{equation}
\begin{gathered}
    \frac{d}{dt}\boldsymbol{q}=\frac{1}{2}\boldsymbol{q}\otimes\begin{bmatrix}0\\\boldsymbol\omega^b\end{bmatrix}\\
    \boldsymbol{M}\frac{d}{dt}\boldsymbol{\nu} + \boldsymbol{C}( \boldsymbol{\nu}) \boldsymbol{\nu} + \boldsymbol{D}( \boldsymbol{\nu}) \boldsymbol{\nu} + \boldsymbol{g}(\boldsymbol{\boldsymbol{q})} =  \boldsymbol{Ka}^b
\end{gathered}
\end{equation}
where $\otimes$ denotes the Hamiltonian product operator, $\boldsymbol{M} \in \mathbb{R}^{6 \times 6}$ represents the mass and inertia, $\boldsymbol{C}( \boldsymbol{\nu})  \in \mathbb{R}^{6 \times 6}$ contains centripetal and Coriolis terms, and $\boldsymbol{D}( \boldsymbol{\nu})  \in \mathbb{R}^{6 \times 6}$ contains the damping terms, while $\boldsymbol{g}(\boldsymbol{q})$ contains the hydrostatic forces and moments, and $\boldsymbol{K} \in \mathbb{R}^{6 \times 6}$ is the thrust gain matrix. 
Then, the objective is to generate actions $\boldsymbol{a}^b = [
a_u, a_v, a_w, a_p, a_q, a_r] \in \mathbb{R}^6$ 
such that 
\begin{equation}
    \begin{aligned}
        \lim_{t\rightarrow\infty} \boldsymbol{v}^b_e(t)&= \boldsymbol{0}\\
        \lim_{t\rightarrow\infty}\boldsymbol{q}_e(t)&= \begin{bmatrix} 1\\\boldsymbol{0}\end{bmatrix}\:.
    \end{aligned}
\end{equation}
In more simple terms, the core objective is to force the linear velocity error to converge to zero and the quaternion error to converge to the identity quaternion.






\section{Proposed Method}\label{sec:method}

\begin{figure}
    \centering
    \includegraphics[width=0.9\linewidth]{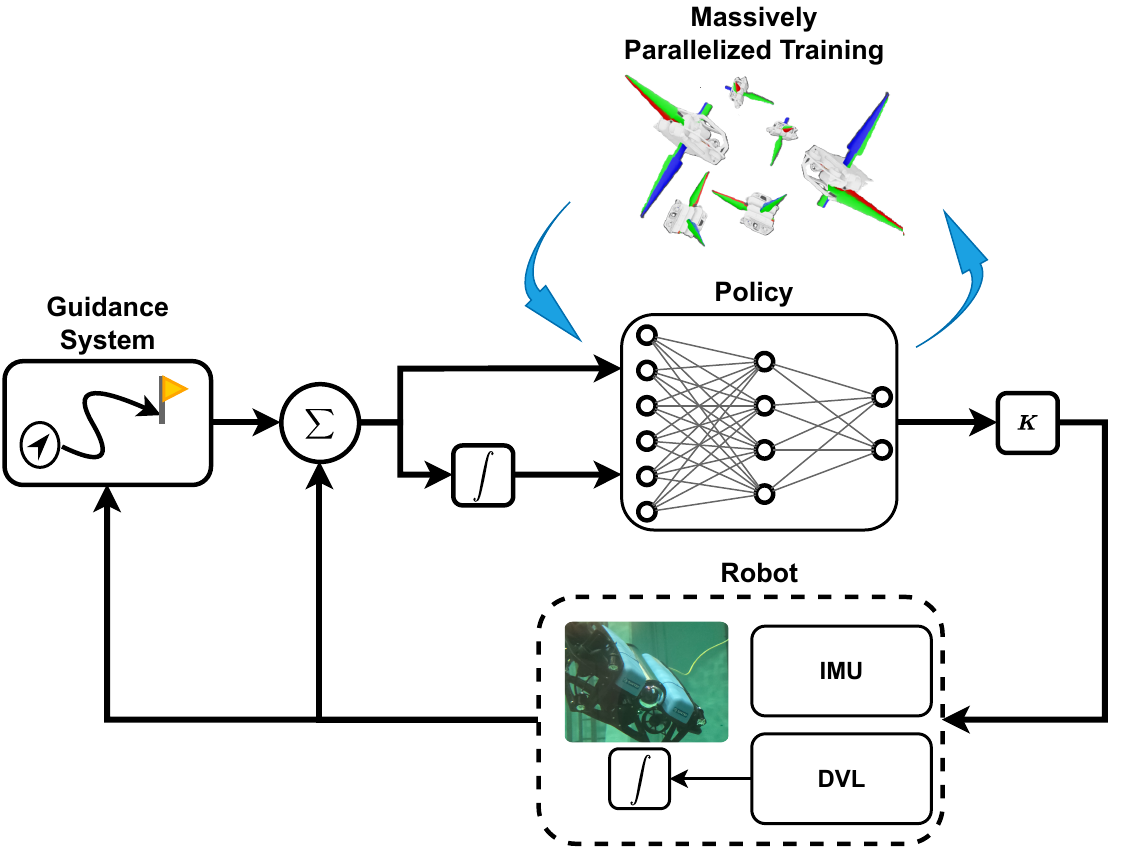}
    \caption{Sim2Swim: The proposed method receives desired and measured linear velocities and orientation, and computes forces and torques, which are sent to the robots thrust allocation scheme.}
    \label{fig:sim2swim}
\end{figure}

The pipeline proposed in this work, depicted in Figure \ref{fig:sim2swim}, consists of a learning environment in Isaac Lab which leverages massive parallelization to generate a general policy for control of velocity and orientation, integral observations in linear velocities and orientation to mitigate steady-state error, and a guidance system that generates desired linear velocities and orientations. 

\subsection{Observation Modeling}
The observation vector $\boldsymbol{o} \in \mathbb{R}^{16}$ consists of the quaternion error, linear velocity errors, and angular velocities. To ensure convergence of the linear velocity and attitude, we also include their integral states.
\begin{equation}
    \boldsymbol{o} = [\boldsymbol{q}_e, \boldsymbol{v}^b_e, \boldsymbol{\omega}^b, \boldsymbol{z}_v, \boldsymbol{z}_q]\:,
\end{equation}
where $\boldsymbol{z}_v\in\mathbb{R}^3$ and $\boldsymbol{z}_q\in\mathbb{R}^3$ represent the integral states of linear velocity error and vector component of the quaternion error, respectively.
Even though, there are no rewards associated with the integral states in the observation vector, they serve to provide the policy with a memory of past observations.

\subsection{Massively Parallelized DRL}
The policy is realized as a 2-layer multilayer perceptron (MLP) with, and is trained with the RSL-RL implementation of the PPO \citep{schulman_proximal_2017} algorithm, readily implemented in Isaac Lab \citep{rudin_learning_2022}.

The action space $\boldsymbol{a}^b \in \mathbb{R}^6$, where each element $a_{\_} \in [-1,1]$ represents the scaled control forces and torques, translates to actual control forces and torques, $\boldsymbol{\tau} \in \mathbb{R}^6$ through
\begin{equation}
    \boldsymbol{\tau} = \boldsymbol{Ka}^b,
\end{equation}
where $\boldsymbol{K}$ is a thrust gain matrix representing the AUV's maximum force or torque in each degree of freedom. 
With this approach, as opposed to directly controlling each thruster with the trained policy, the RL algorithm does not need to learn a thrust allocation scheme; an already solved problem and trivially calculated problem for AUVs~\citep{Johansen:2013:automatica:control_allocation}. 
Work such as \cite{eschmann_learning_2024} argues in favor of a low-level policy that outputs motor commands. However, the motor time-constant plays a larger role for nanocopters since these usually have a high thrust-to-weight ratio. Additionally, unlike our approach, this approach puts an assumption on the number of thrusters used and therefore makes the policy design specific to each vessel. 

\subsection{Reward Formulation}
The reward function is formulated as the sum 
\begin{equation}
r = \sum_i r_i + r_q + r_a,
\end{equation}
where each term $r_i$ represents the reward associated with each set of observations $o_i \in \{ \boldsymbol{q}_e, \boldsymbol{v}^b_e, \boldsymbol{\omega}^b \}$ is formulated as
\begin{equation}
    r_i = w_i e^{- \Vert o_i \Vert^2},
\end{equation}
where $w_{\_}$ is the associated weight.
The reward for the attitude error is defined as
\begin{equation}
    r_q =  w_q e^{- \angle (\boldsymbol{q}_d, \boldsymbol{q})},
\end{equation}
where $\angle (\boldsymbol{q}_d, \boldsymbol{q})$ signify the rotation difference between $\boldsymbol{q}_d$ and $\boldsymbol{q}$.
Additionally, we add a reward 
\begin{equation}
    r_a = w_a e^{- \Vert \boldsymbol{a} \Vert}
\end{equation}
to minimize the actions taken.

\subsection{Domain Randomization}
To achieve a policy that is robust to parametric uncertainties, we employ a similar domain randomization as in \cite{cai_learning_2025}.
The mass and volume of the robot are varied with uniformly sampling, while the offset between the center of buoyancy (CB) with the center of mass (CM) is uniformly sampled in a sphere.

\subsection{Desired States}
In each training episode each individual AUV is given a time-varying desired orientation with random initial conditions that follows the Frenet-Serret frame of a trajectory with its velocities defined as
\begin{equation}
    \boldsymbol{v}(t) = [a , b \sin(\omega t), c \cos (\omega t)].
\end{equation}
Finally, the desired body velocities are randomly sampled for each episode $\boldsymbol{v}^b_d(t)$ with speed $||\boldsymbol{v}^b_d(t)||=V_d = 0.5$ \si{m/s} on the unit sphere, to variate the direction.

\section{Experimental Validation}\label{sec:sim2real}
\subsection{Hardware and Training Setup}
\begin{figure}[h]
    \centering
    \includegraphics[width=0.7\linewidth]{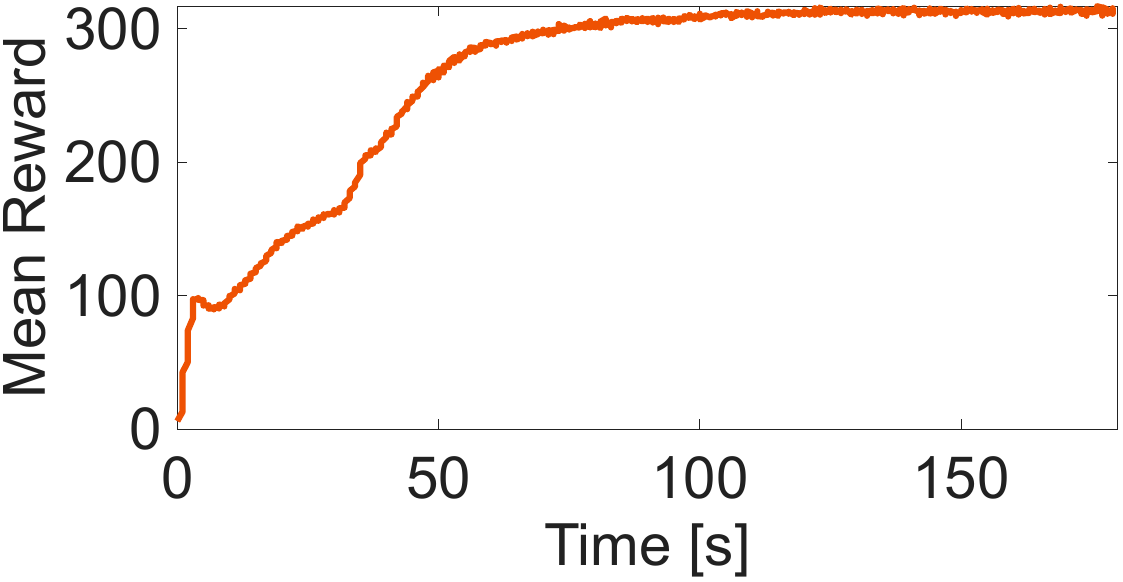}
    \caption{Mean reward against training time.}
    \label{fig:mean_reward_time}
\end{figure}

The policy is trained on a PC equipped with an Intel Core i7-12800HX CPU, Nvidia A2000 GPU with 8GB of VRAM, and 32GB of RAM. The maximum episode length is set to 5 seconds, with 2048 parallel learning environments. Table \ref{tab:train:hyper_parameters} contains the training weights and parameters used during training. The training is completed in less than 3 minutes. In Figure \ref{fig:mean_reward_time}, we see that the policy converges after about 80 seconds, with a mean reward in the final learning iteration of 315.

\begin{table}[h]
    \centering
    \caption{Training weights and parameters}
    \begin{tabular}{lcc}
        Parameter & Symbol  & Value \\ \hline
        Orientation error & $w_{q}$ & 0.4 \\ 
       Angular velocity & $w_\omega$ & 0.05 \\
       Linear velocity & $w_v$ & 0.2 \\
        Actions & $w_{a}$ & 0.3 \\
        Trajectory frequency & $\omega$ & 0.2 \\
        Trajectory coefficients  & [$a$, $b$, $c$]& [0.5 0.5 0.3] \\
    \end{tabular}
    \label{tab:train:hyper_parameters}
\end{table}

\subsection{Sim2Real Transfer}

\begin{figure*}[t!]
    \centering
\begin{subfigure}[t]{0.32\textwidth}
    \centering
    \includegraphics[width=\linewidth]{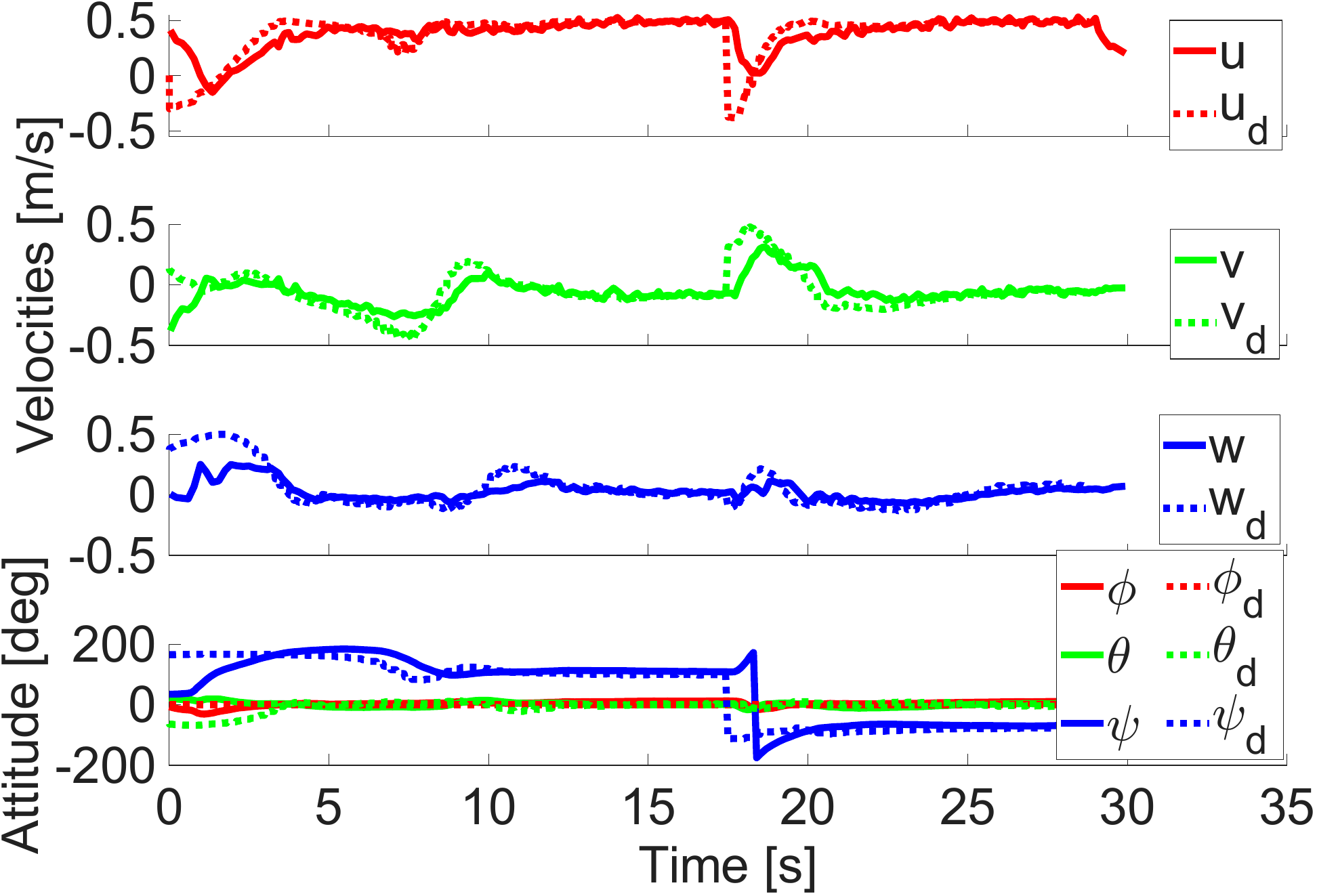}
\end{subfigure}
\begin{subfigure}[t]{0.32\textwidth}
    \centering
    \includegraphics[width=\linewidth]{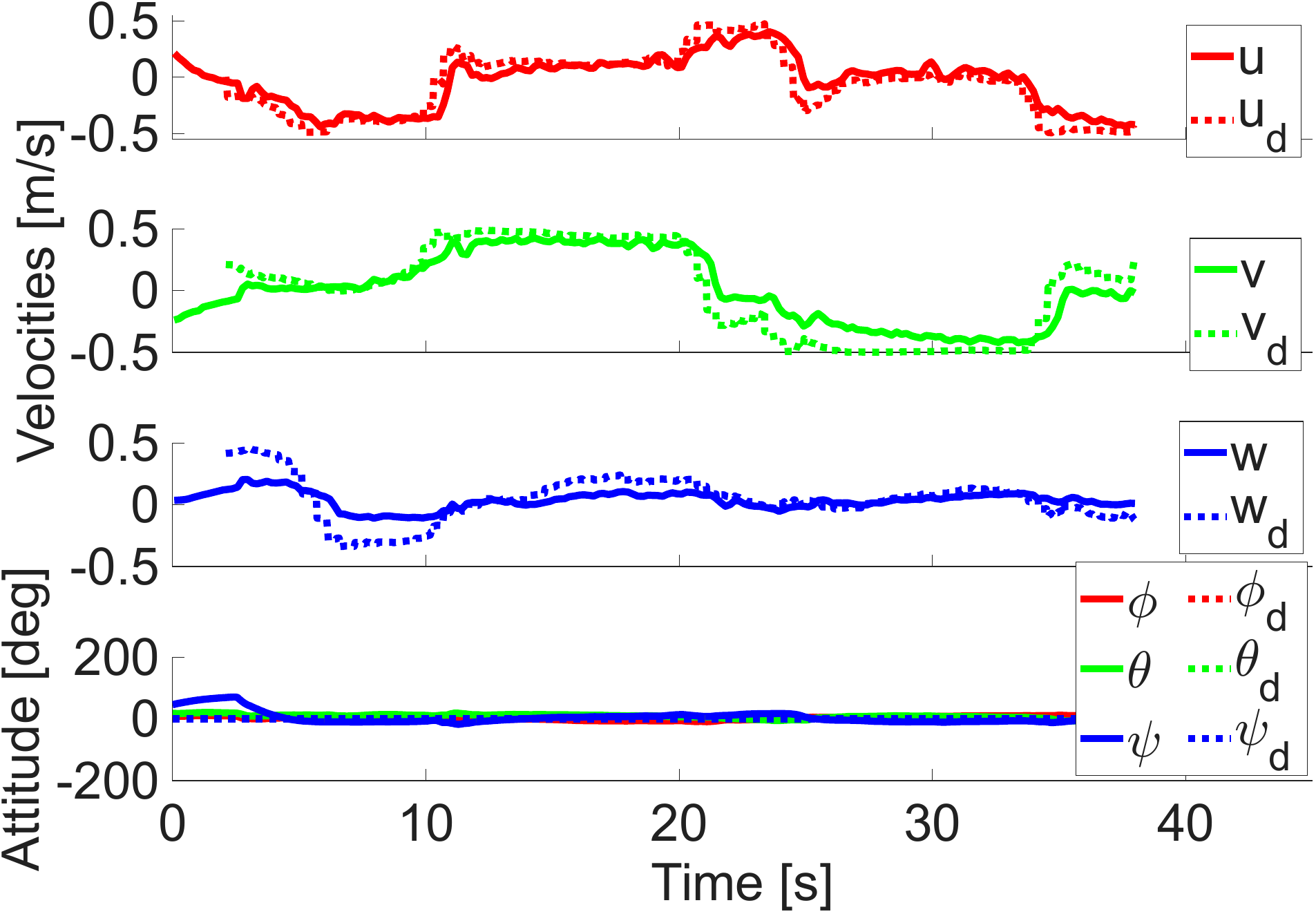}
\end{subfigure}
\begin{subfigure}[t]{0.32\textwidth}
    \centering
    \includegraphics[width=\linewidth]{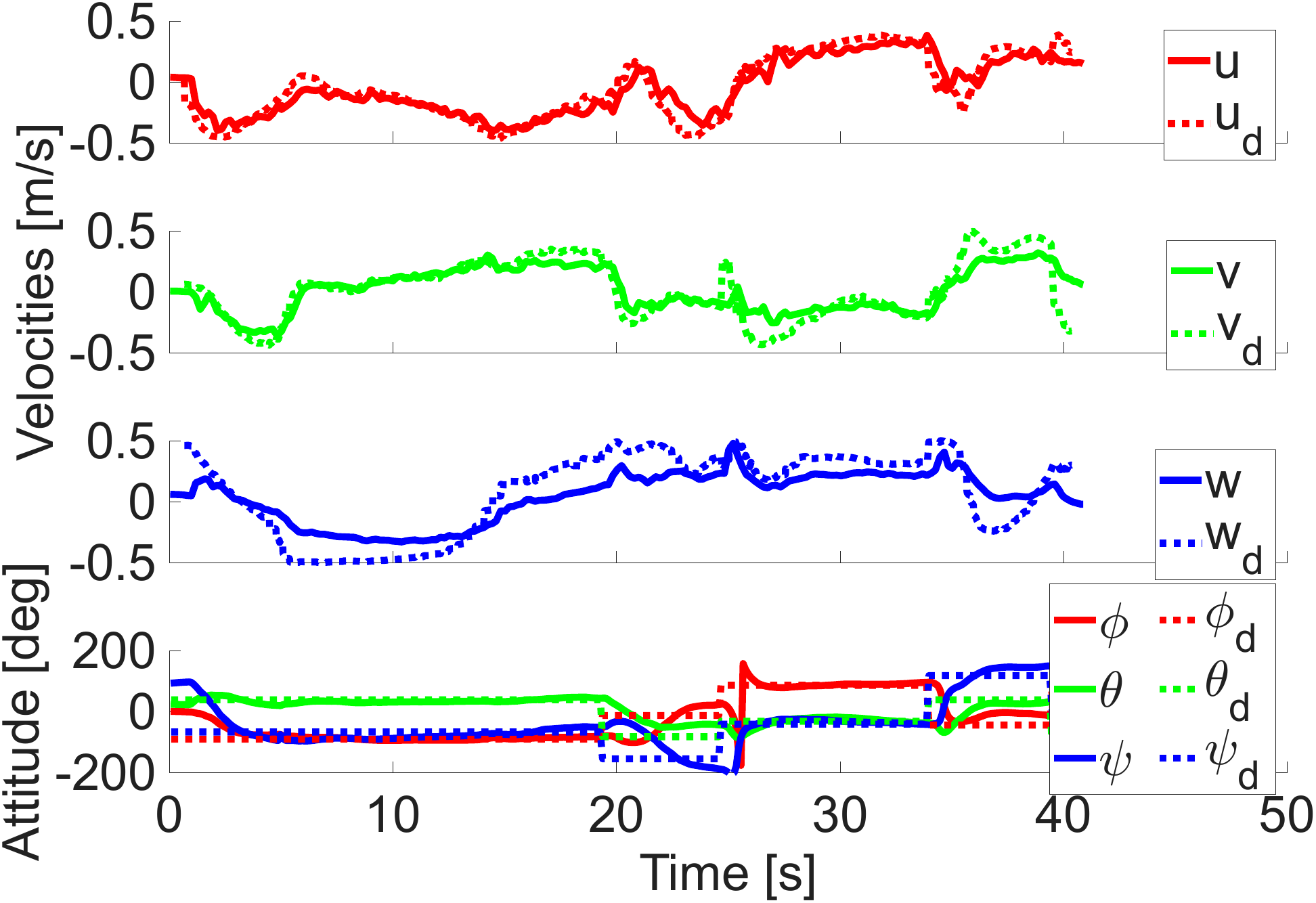}
\end{subfigure}
\begin{subfigure}[t]{0.32\textwidth}
    \centering
    \includegraphics[width=\linewidth]{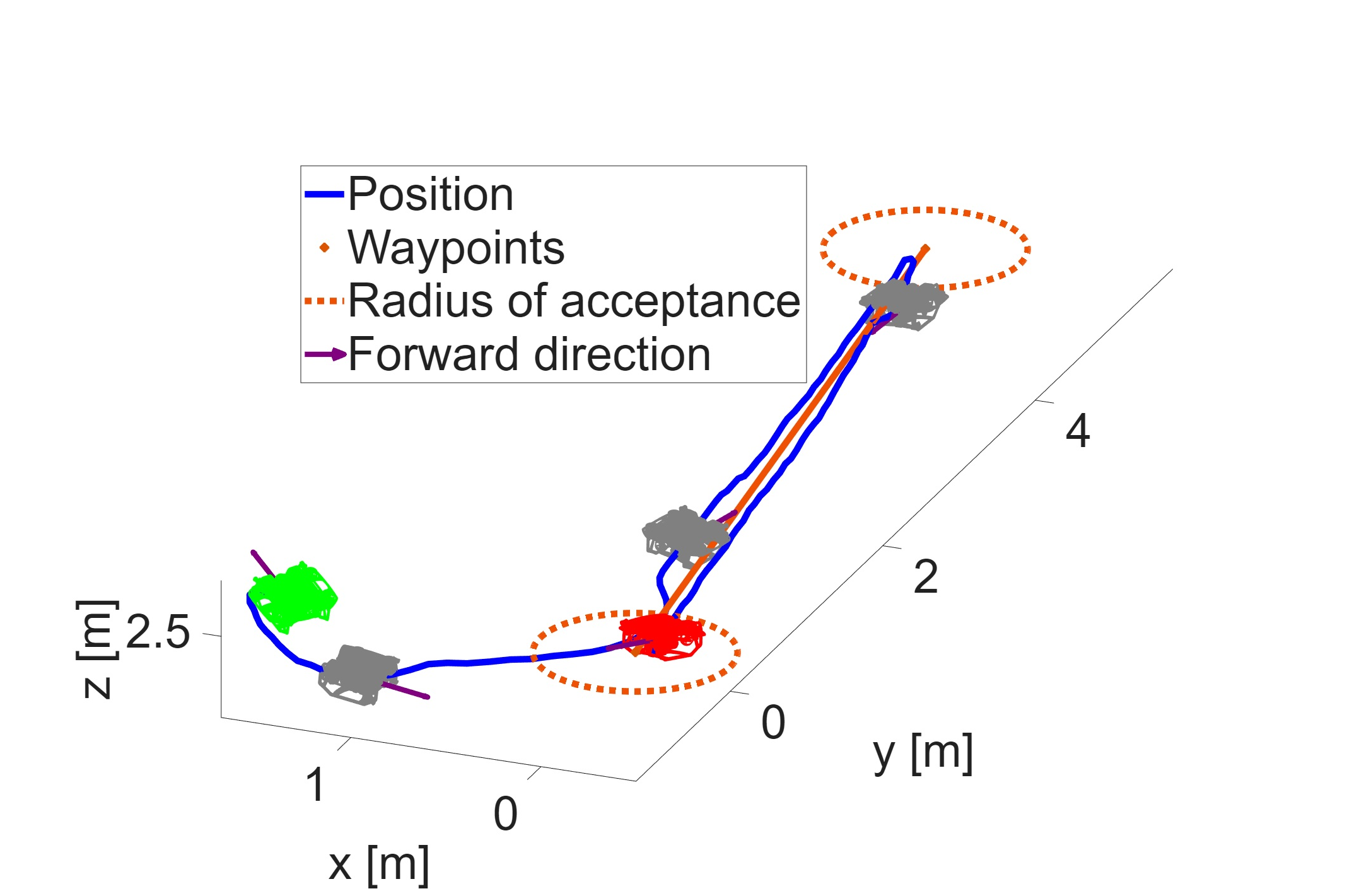}
    \caption{Straight-line}
    \label{fig:exp:straightline}
\end{subfigure}
\begin{subfigure}[t]{0.32\textwidth}
    \centering
    \includegraphics[width=\linewidth]{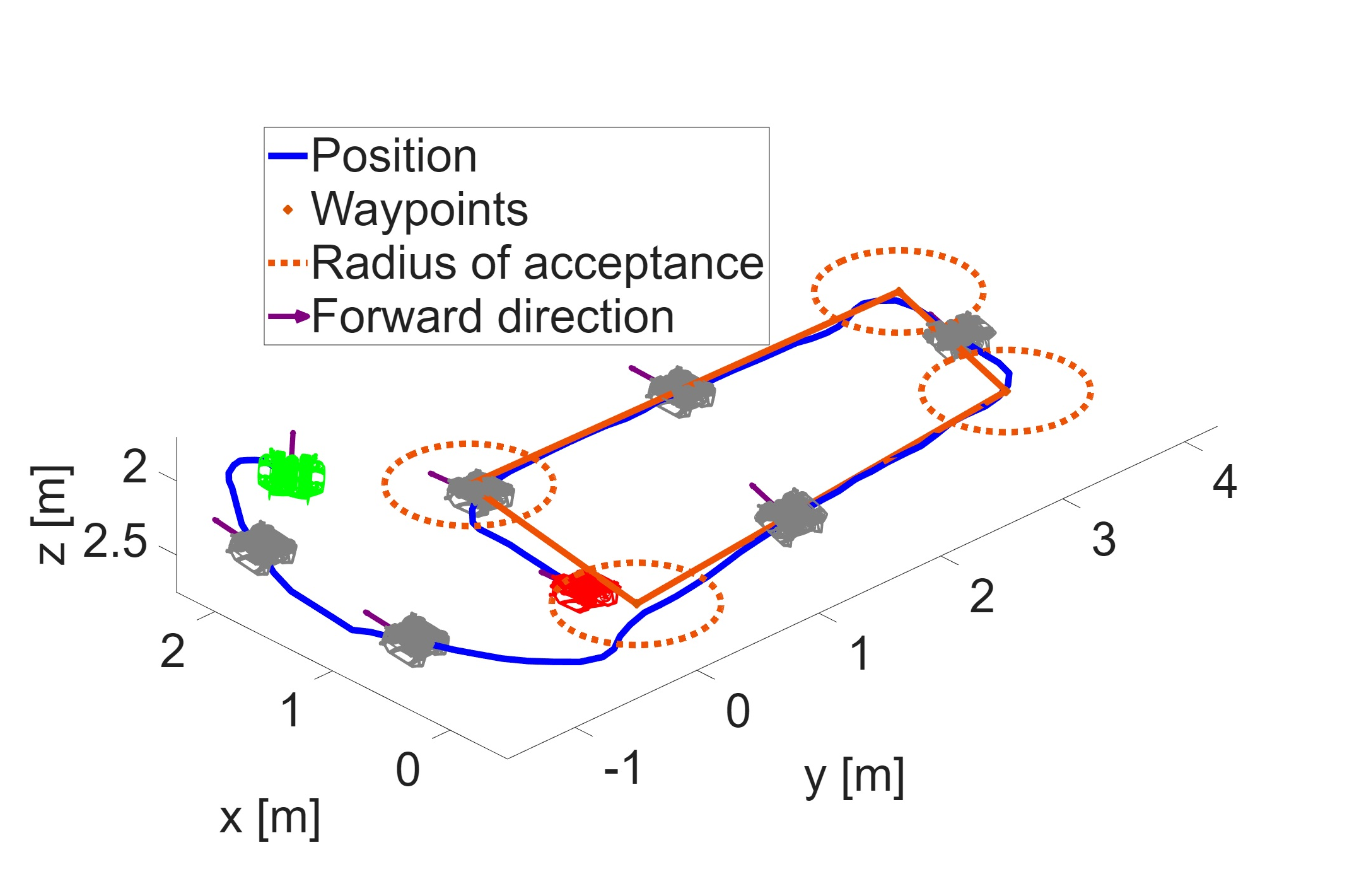}
    \caption{Ballast}
    \label{fig:exp:ballast}
\end{subfigure}
\begin{subfigure}[t]{0.32\textwidth}
    \centering
    \includegraphics[width=\linewidth]{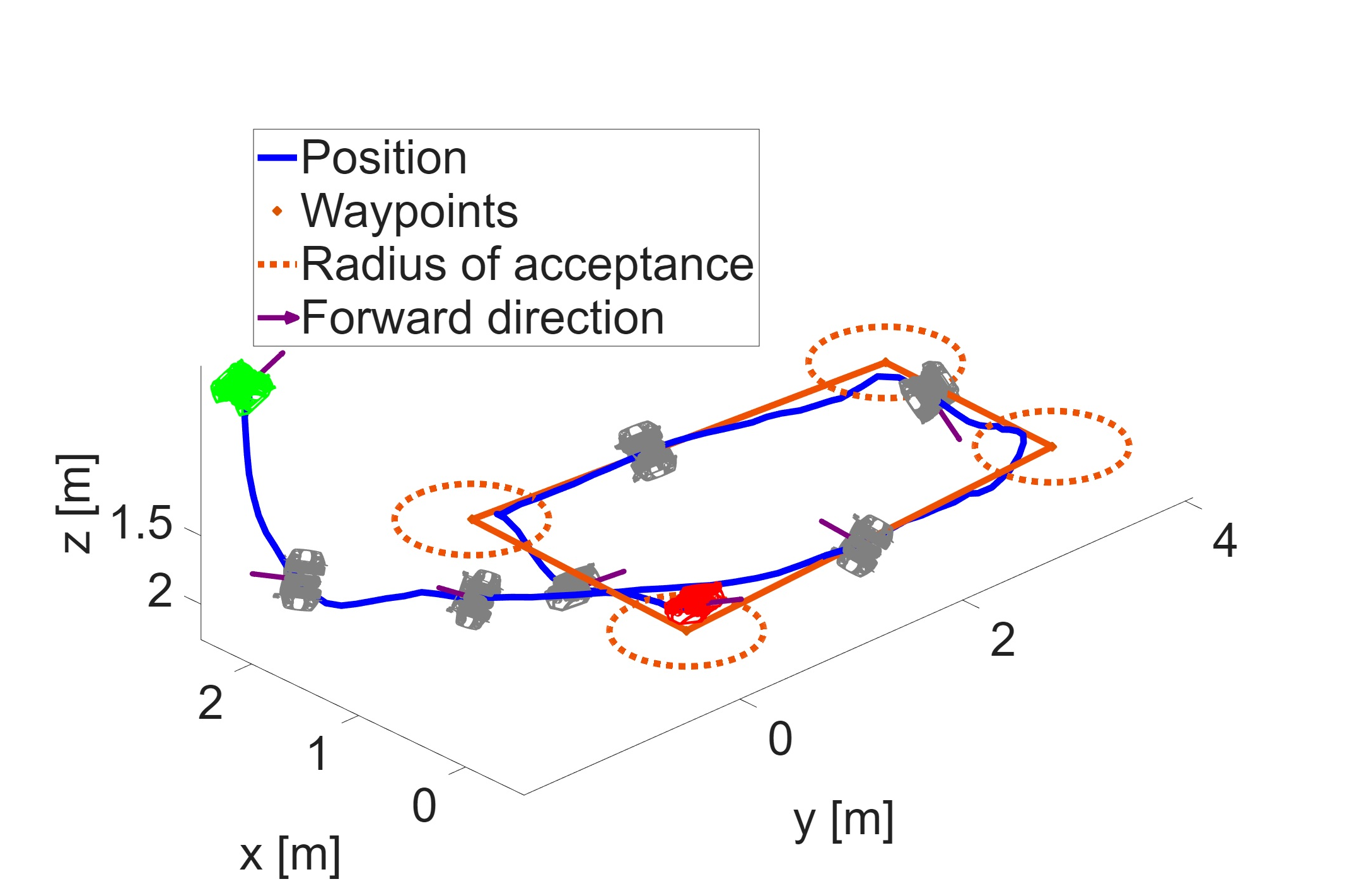}
    \caption{Random orientation}
    \label{fig:exp:rand_orint}
\end{subfigure}
\caption{Experiment results of three separate trials. Left column shows the ROV following a straight line, where the desired heading and pitch directs the ROV toward the velocity direction. In the middle column, a ballast is attached to the port-side of the ROV. Here, the ROV following a rectangular path, with a constant heading setpoint. In the right column, the ROV follows a square path, with random desired orientations given at each waypoint. 
Snapshots of the ROV are included in the lower row, with green and red  signifying the starting and final position, respectively.}
\label{fig:experiments}
\end{figure*}

We validate the policy on a BlueRobotics BlueROV2 Heavy, depicted in Figures~\ref{fig:murk:b} and \ref{fig:murk:c} in an indoor pool.
The vehicle is equipped with a Water Linked A50 Doppler velocity log measuring body velocity $\boldsymbol{v}^b$ and estimating position ($x$ and $y$). 
We measure the depth ($z$, positive down) with a BlueRobotics Bar30 pressure sensor,
while the orientation is provided by the inertial navigation system of the BlueROV2.
We employ a 3D line-of-sight (LOS) guidance law~\citep{Breivik:2005:CDC} to generate linear velocity reference signals that ensures vehicle converge to the path, independent of its attitude. 
This leaves the rotational degrees of freedom available to simultaneously perform any motion or obtain any orientation.

We present results from a set of three separate trials, with the associated results reported in Figure~\ref{fig:experiments}. 
Each trial is designed to assess a specific aspect of controller performance. 
First, the vehicle is commanded to follow a straight line back and forth, with desired heading and pitch equal to the desired course and elevation angles calculated by the LOS guidance law. 
In the second trial, we add a 600\si{g} ballast load to the port side of the vehicle to assess policy robustness under changes in mass and offsets to the CM, and command the vehicle to reach four waypoints arranged in a square. 
This load represents a $5\%$ increase in mass and changes the vehicle's buoyancy from positive to negative. 
In the final test, the vehicle is again commanded to reach a set of four waypoints in a square, but this time the desired orientation is given by setpoints randomly generated and changed at each waypoint ($\phi_i\sim\mathcal{U}(-\frac{\pi}{2},\frac{\pi}{2})$, $\theta_i\sim\mathcal{U}(-\frac{\pi}{2},\frac{\pi}{2})$, $\psi_i\sim\mathcal{U}[-\pi,\pi], i\in\{1,2,3,4\}$ for roll, pitch, and yaw angles).
We do this to demonstrate that the policy can achieve accurate path following and waypoint convergence even with unconventional attitudes that ---unlike common practice --- are not dictated by the path. 
Tracing a path while holding such configurations can enable new and improved inspection capabilities for applications where the area of interest is not parallel to the path.

 In Figure \ref{fig:exp:straightline}, where the vehicle traces a straight line path back and forth, we see that the linear velocity components converge to their desired values with no steady-state errors. The heading angle ($\psi$) converges smoothly to its desired value ($\psi_d$) without steady-state error, while the remaining Euler angles ($\phi,\theta$) remains close to zero.  

In Figure \ref{fig:exp:ballast}, the vehicle is equipped with a ballast and traces a square path while the desired Euler angles are set to zero. The vehicle closely tracks the desired surge velocity. In sway and heave, it can be noted that the vehicle is unable to track fast variations, but it converges to the desired values. 
For the Euler angles, some offsets when the desired linear velocities change abruptly can be seen.
Most likely, this is due to imperfect thrust allocation, which generates a small moment on the vehicle. 
The policy shows no significant degradation and appears robust to parametric uncertainties in mass and center of mass.

In Figure \ref{fig:exp:rand_orint}, similarly to previous trials, the vehicle is able to track linear velocities in surge and sway. We observe a slower response in heave, which is also seen in Figure \ref{fig:exp:ballast}. The effect of this can be observed in the 3D plot of the path, where the inability to track the large dip in desired heave 37 seconds into the experiment causes the depth to deviate slightly from the desired path. This could be caused by a loss of thrust in the heave DOF due to the vehicle's unconventional orientations, which results in sub-optimal thruster utilization. We see that the vehicle is able to hold its attitude even at extreme pitch and roll angles.

\section{Conclusion}\label{sec:conclusion}
This paper presented Sim2Swim, the first deep reinforcement learning-based controller capable of agile underwater path-following in 6 DOF, trained in less than 3 minutes. 
Through extensive experimental validation, the policy showcased robustness to parametric uncertainties and was able to track both linear velocities and attitude, including extreme roll and pitch angles, enabling agile path following and maneuvering.
We strongly believe that this work serves as the foundation to enable new advanced inspection behaviors of complex subsea infrastructure.
Future work will focus on validation in exposed underwater conditions, such as in aquaculture and wind-farm inspections, as well as extensions to non-holonomic systems.

\begin{ack}
This work is funded by the European Union through the INESCTEC.OCEAN Center of Excellence in Ocean Research and Engineering (Project number 101136903). 
\end{ack}

\bibliography{ifacconf}             

@article{diamanti2025marine,
  title={Marine Archaeological Surveying Using Snake Robots: The Eely Survey of Figaro Wreck in the High Arctic},
  author={Diamanti, Eleni and Fossdal, Markus and Iversflaten, Markus H{\o}gevoll and S{\ae}b{\o}, Bj{\o}rn K{\aa}re and Kasparavi{\v{c}}i{\=u}t{\.e}, Gabriel{\.e} and Waldum, Ambj{\o}rn Grimsrud and Yip, Mauhing and {\O}deg{\aa}rd, {\O}yvind and De La Torre, Pedro and Pettersen, Kristin Y and others},
  journal={Marine Technology Society Journal},
  volume={59},
  number={2},
  pages={78--103},
  year={2025},
  publisher={Marine Technology Society}
}

@article{eschmann_learning_2024,
	title = {Learning to {Fly} in {Seconds}},
	volume = {9},
	issn = {2377-3766},
	url = {https://ieeexplore.ieee.org/document/10517383/},
	doi = {10.1109/LRA.2024.3396025},
	number = {7},
	urldate = {2025-11-24},
	journal = {IEEE Robotics and Automation Letters},
	author = {Eschmann, Jonas and Albani, Dario and Loianno, Giuseppe},
	month = jul,
	year = {2024},
	pages = {6336--6343},
}

@article{sufan_swim4real_2025,
	title = {{Swim4Real}: {Deep} {Reinforcement} {Learning}-{Based} {Energy}-{Efficient} and {Agile} 6-{DOF} {Control} for {Underwater} {Vehicles}},
	volume = {10},
	issn = {2377-3766},
	shorttitle = {{Swim4Real}},
	url = {https://ieeexplore.ieee.org/abstract/document/11020757},
	doi = {10.1109/LRA.2025.3575650},
	number = {7},
	urldate = {2025-11-19},
	journal = {IEEE Robotics and Automation Letters},
	author = {Sufán, Vicente and Troni, Giancarlo},
	month = jul,
	year = {2025},
	pages = {7326--7333},
}

@article{ma_neural_2024,
	title = {Neural {Network} {Model}-{Based} {Reinforcement} {Learning} {Control} for {AUV} 3-{D} {Path} {Following}},
	volume = {9},
	issn = {2379-8904},
	url = {https://ieeexplore.ieee.org/abstract/document/10143677},
	doi = {10.1109/TIV.2023.3282681},
	number = {1},
	urldate = {2025-11-12},
	journal = {IEEE Transactions on Intelligent Vehicles},
	author = {Ma, Dongfang and Chen, Xi and Ma, Weihao and Zheng, Huarong and Qu, Fengzhong},
	month = jan,
	year = {2024},
	pages = {893--904},
}

@article{wang_imitation_2025,
	title = {Imitation learning from observation for {ROV} path tracking},
	volume = {3},
	issn = {2948-1953},
	url = {https://doi.org/10.1007/s44295-025-00069-0},
	doi = {10.1007/s44295-025-00069-0},
	language = {en},
	number = {1},
	urldate = {2025-11-11},
	journal = {Intelligent Marine Technology and Systems},
	author = {Wang, Jun and Xiang, Song and Shen, Tian and Fang, Zheng and Niu, Shilong and Pan, Xingwei and Li, Guangliang},
	month = jul,
	year = {2025},
	pages = {20},
}

@inproceedings{herland_non-prehensile_2025,
	title = {Non-{Prehensile} {Shape} {Manipulation} of {Elastoplastic} {Objects} {With} {Reinforcement} {Learning}},
	url = {https://ieeexplore.ieee.org/document/11127639/},
	doi = {10.1109/ICRA55743.2025.11127639},
	urldate = {2025-11-11},
	booktitle = {2025 {IEEE} {International} {Conference} on {Robotics} and {Automation} ({ICRA})},
	author = {Herland, Sverre and Misimi, Ekrem},
	month = may,
	year = {2025},
	pages = {13204--13210},
}

@article{tsounis_deepgait_2020,
	title = {{DeepGait}: {Planning} and {Control} of {Quadrupedal} {Gaits} {Using} {Deep} {Reinforcement} {Learning}},
	volume = {5},
	issn = {2377-3766},
	shorttitle = {{DeepGait}},
	url = {https://ieeexplore.ieee.org/abstract/document/9028188},
	doi = {10.1109/LRA.2020.2979660},
	number = {2},
	urldate = {2025-11-11},
	journal = {IEEE Robotics and Automation Letters},
	author = {Tsounis, Vassilios and Alge, Mitja and Lee, Joonho and Farshidian, Farbod and Hutter, Marco},
	month = apr,
	year = {2020},
	pages = {3699--3706},
}

@article{hadi_deep_2022,
	title = {Deep reinforcement learning for adaptive path planning and control of an autonomous underwater vehicle},
	volume = {129},
	issn = {0141-1187},
	url = {https://www.sciencedirect.com/science/article/pii/S0141118722002589},
	doi = {10.1016/j.apor.2022.103326},
	urldate = {2025-11-07},
	journal = {Applied Ocean Research},
	author = {Hadi, Behnaz and Khosravi, Alireza and Sarhadi, Pouria},
	month = dec,
	year = {2022},
	pages = {103326},
}

@inproceedings{gu_deep_2017,
	title = {Deep reinforcement learning for robotic manipulation with asynchronous off-policy updates},
	url = {https://ieeexplore.ieee.org/abstract/document/7989385},
	doi = {10.1109/ICRA.2017.7989385},
	urldate = {2025-11-07},
	booktitle = {2017 {IEEE} {International} {Conference} on {Robotics} and {Automation} ({ICRA})},
	author = {Gu, Shixiang and Holly, Ethan and Lillicrap, Timothy and Levine, Sergey},
	month = may,
	year = {2017},
	pages = {3389--3396},
}

@inproceedings{panerati_learning_2021,
	title = {Learning to {Fly}—a {Gym} {Environment} with {PyBullet} {Physics} for {Reinforcement} {Learning} of {Multi}-agent {Quadcopter} {Control}},
	url = {https://ieeexplore.ieee.org/abstract/document/9635857},
	doi = {10.1109/IROS51168.2021.9635857},
	urldate = {2025-11-07},
	booktitle = {2021 {IEEE}/{RSJ} {International} {Conference} on {Intelligent} {Robots} and {Systems} ({IROS})},
	author = {Panerati, Jacopo and Zheng, Hehui and Zhou, SiQi and Xu, James and Prorok, Amanda and Schoellig, Angela P.},
	month = sep,
	year = {2021},
	note = {ISSN: 2153-0866},
	pages = {7512--7519},
}

@misc{schulman_proximal_2017,
	title = {Proximal {Policy} {Optimization} {Algorithms}},
	url = {http://arxiv.org/abs/1707.06347},
	doi = {10.48550/arXiv.1707.06347},
	urldate = {2025-09-24},
	publisher = {arXiv},
	author = {Schulman, John and Wolski, Filip and Dhariwal, Prafulla and Radford, Alec and Klimov, Oleg},
	month = aug,
	year = {2017},
	note = {arXiv:1707.06347 [cs]},
}

@misc{rudin_learning_2022,
	title = {Learning to {Walk} in {Minutes} {Using} {Massively} {Parallel} {Deep} {Reinforcement} {Learning}},
	url = {http://arxiv.org/abs/2109.11978},
	doi = {10.48550/arXiv.2109.11978},
	urldate = {2025-09-24},
	publisher = {arXiv},
	author = {Rudin, Nikita and Hoeller, David and Reist, Philipp and Hutter, Marco},
	month = aug,
	year = {2022},
	note = {arXiv:2109.11978 [cs]},
}

@inproceedings{cai_learning_2025,
	title = {Learning to {Swim}: {Reinforcement} {Learning} for 6-{DOF} {Control} of {Thruster}-{Driven} {Autonomous} {Underwater} {Vehicles}},
	shorttitle = {Learning to {Swim}},
	url = {https://ieeexplore.ieee.org/document/11128688/},
	doi = {10.1109/ICRA55743.2025.11128688},
	urldate = {2025-09-23},
	booktitle = {2025 {IEEE} {International} {Conference} on {Robotics} and {Automation} ({ICRA})},
	author = {Cai, Levi and Chang, Kevin and Girdhar, Yogesh},
	month = may,
	year = {2025},
	pages = {11286--11293},
}

@article{ohrem_application_2024,
	title = {Application of modified {Model} {Reference} {Adaptive} {Controller} and {Observer} ({MRACO}) for speed control of an unmanned underwater vehicle},
	volume = {58},
	issn = {24058963},
	url = {https://linkinghub.elsevier.com/retrieve/pii/S2405896324018093},
	doi = {10.1016/j.ifacol.2024.10.054},
	language = {en},
	number = {20},
	urldate = {2025-09-05},
	journal = {IFAC-PapersOnLine},
	author = {Ohrem, Sveinung Johan and Haugaløkken, Bent Oddvar Arnesen and Holden, Christian},
	year = {2024},
	pages = {196--202},
}

@book{fossen_handbook_2021,
	edition = {2nd},
	title = {Handbook of {Marine} {Craft} {Hydrodynamics} and {Control}},
	isbn = {978-1-119-57503-0},
	language = {English},
	publisher = {John Wiley \& Sons},
	author = {Fossen, Thor I.},
	year = {2021},
}

@article{carlucho2018adaptive,
  title={Adaptive low-level control of autonomous underwater vehicles using deep reinforcement learning},
  author={Carlucho, Ignacio and De Paula, Mariano and Wang, Sen and Petillot, Yvan and Acosta, Gerardo G},
  journal={Robotics and Autonomous Systems},
  volume={107},
  pages={71--86},
  year={2018},
  publisher={Elsevier}
}

@article{caharija2016integral,
  title={Integral line-of-sight guidance and control of underactuated marine vehicles: Theory, simulations, and experiments},
  author={Caharija, Walter and Pettersen, Kristin Y and Bibuli, Marco and Calado, Pedro and Zereik, Enrica and Braga, Jos{\'e} and Gravdahl, Jan Tommy and S{\o}rensen, Asgeir J and Milovanovi{\'c}, Milan and Bruzzone, Gabriele},
  journal={IEEE Transactions on Control Systems Technology},
  volume={24},
  number={5},
  pages={1623--1642},
  year={2016},
  publisher={IEEE}
}

@article{fossum2019toward,
  title={Toward adaptive robotic sampling of phytoplankton in the coastal ocean},
  author={Fossum, Trygve O and Fragoso, Glaucia M and Davies, Emlyn J and Ullgren, Jenny E and Mendes, Renato and Johnsen, Geir and Ellingsen, Ingrid and Eidsvik, Jo and Ludvigsen, Martin and Rajan, Kanna},
  journal={Science Robotics},
  volume={4},
  number={27},
  year={2019},
  publisher={American Association for the Advancement of Science}
}

@article{khalid2022applications,
  title={Applications of robotics in floating offshore wind farm operations and maintenance: Literature review and trends},
  author={Khalid, Omer and Hao, Guangbo and Desmond, Cian and Macdonald, Hamish and McAuliffe, Fiona Devoy and Dooly, Gerard and Hu, Weifei},
  journal={Wind Energy},
  volume={25},
  number={11},
  pages={1880--1899},
  year={2022},
  publisher={Wiley Online Library}
}

@incollection{kelasidi2023robotics,
  title={Robotics for sea-based fish farming},
  author={Kelasidi, Eleni and Svendsen, Eirik},
  booktitle={Encyclopedia of smart agriculture technologies},
  pages={1--20},
  year={2023},
  publisher={Springer}
}

@inproceedings{transeth2024safesub,
  title={SAFESUB: Safe and Autonomous Subsea Intervention},
  author={Transeth, Aksel A and Thorstensen, Jostein and Mohammed, Ahmed and Thielemann, Jens T and Ening, Klaus and Gr{\o}tli, Esten Ingar and Haugal{\o}kken, Bent OA and Brandt, Martin Albertsen and M{\o}ller, Ments Tore and Hovland, Rebecca Petterteig and others},
  booktitle={2024 20th IEEE/ASME International Conference on Mechatronic and Embedded Systems and Applications (MESA)},
  pages={1--8},
  year={2024},
  organization={IEEE}
}

@article{Ludvigsen:2016:autonomous_mapping,
title = {Towards integrated autonomous underwater operations for ocean mapping and monitoring},
journal = {Annual Reviews in Control},
volume = {42},
pages = {145-157},
year = {2016},
issn = {1367-5788},
doi = {https://doi.org/10.1016/j.arcontrol.2016.09.013},
url = {https://www.sciencedirect.com/science/article/pii/S1367578816300256},
author = {Martin Ludvigsen and Asgeir J. Sørensen}
}

@INPROCEEDINGS{Breivik:2005:CDC,
  author={Breivik, M. and Fossen, T.I.},
  booktitle={Proc. 44th IEEE Conference on Decision and Control}, 
  title={Principles of Guidance-Based Path Following in {2D} and {3D}}, 
  year={2005},
  volume={},
  number={},
  pages={627-634},
  doi={10.1109/CDC.2005.1582226}}

@article{Shtessel:2012:automatica:supertwisting,
title = {A novel adaptive-gain supertwisting sliding mode controller: Methodology and application},
journal = {Automatica},
volume = {48},
number = {5},
pages = {759-769},
year = {2012},
issn = {0005-1098},
doi = {https://doi.org/10.1016/j.automatica.2012.02.024},
url = {https://www.sciencedirect.com/science/article/pii/S0005109812000751},
author = {Yuri Shtessel and Mohammed Taleb and Franck Plestan}
}

@misc{nvidia:2025:isaac_lab,
    title={Isaac {Lab}: A {GPU}-Accelerated Simulation Framework for Multi-Modal Robot Learning}, 
    author={Mayank Mittal and {et al.}},
    year={2025},
    eprint={2511.04831},
    archivePrefix={arXiv},
    primaryClass={cs.RO},
    url={https://arxiv.org/abs/2511.04831}
}

@article{Johansen:2013:automatica:control_allocation,
title = {Control allocation—A survey},
journal = {Automatica},
volume = {49},
number = {5},
pages = {1087-1103},
year = {2013},
issn = {0005-1098},
doi = {https://doi.org/10.1016/j.automatica.2013.01.035},
url = {https://www.sciencedirect.com/science/article/pii/S0005109813000368},
author = {Tor A. Johansen and Thor I. Fossen}
}

@inproceedings{xanthidis_aquavis_2021,
	title = {{AquaVis}: {A} {Perception}-{Aware} {Autonomous} {Navigation} {Framework} for {Underwater} {Vehicles}},
	shorttitle = {{AquaVis}},
	url = {https://ieeexplore.ieee.org/abstract/document/9636124},
	doi = {10.1109/IROS51168.2021.9636124},
	urldate = {2025-12-04},
	author = {Xanthidis, Marios and Kalaitzakis, Michail and Karapetyan, Nare and Johnson, James and Vitzilaios, Nikolaos and O’Kane, Jason M. and Rekleitis, Ioannis},
	month = sep,
	year = {2021},
	pages = {5410--5417},
    booktitle = {2021 IEEE/RSJ International Conference on Intelligent Robots and Systems (IROS)},
    issn = {123},
    volume = {1},
    
}
                                                   








     
\end{document}